# A Flexible Connector for Soft Modular Robots Based on Micropatterned Intersurface Jamming

Yu Alexander Tse, Shuai Liu, Yang Yang, *Member, IEEE*, and Michael Yu Wang, *Fellow, IEEE*

*Abstract—* Soft modular robots enable more flexibility and safer interaction with the changing environment than traditional robots. However, it has remained challenging to create deformable connectors that can be integrated into soft machines. In this work, we propose a flexible connector for soft modular robots based on micropatterned intersurface jamming. The connector is composed of micropatterned dry adhesives made by silicone rubber and a flexible main body with inflatable chambers for active engagement and disengagement. Through connection force tests, we evaluate the characteristics of the connector both in the linear direction and under rotational disruptions. The connector can stably support an average maximum load of 22 N (83 times the connector's body weight) linearly and 10.86 N under planar rotation. The proposed connector demonstrates the potential to create a robust connection between soft modular robots without raising the system's overall stiffness; thus guarantees high flexibility of the robotic system.

## I. INTRODUCTION

To date, various types of modular robots have been developed to self-reconfigure and adapt to changing environments by rearranging their modules [1]. Compared to conventional fixed-morphology robots, modular robots have the potential to be more functionally flexible and robust [2]. Furthermore, a soft modular robot with variable stiffness cells can display physical compliance enabling safer interaction with the environment [3]. However, to the best of our knowledge, creating soft modular robots without a rigid body connection mechanism remains a significant challenge that limits their practical application.

Most state of art modular robots propose rigid building blocks with incompliant connection mechanisms [4], such as magnets and mechanical latching. Despite the fact that rigid connectors provide stability and controllability to the robotic system, it limits the overall flexibility [5-7]. As a result, modular robots based on the aforementioned connection methods can't adapt efficiently to the changing environment. For example, a rigid modular robot can't go through a confined space smaller than its individual building blocks. Therefore, it is meaningful to build soft modular robots with flexible connection mechanisms. One possible solution to create

*Research supported by ABC Foundation. the Hong Kong Innovation and Technology Fund (ITF) ITS-376-16FP.

Y. Tse, S. Liu and Y. Yang are with the Department of Mechanical and Aerospace Engineering, Hong Kong University of Science and Technology, Hong Kong (e-mail: yatse@connect.ust.hk; sliubw@ust.hk). Y. Yang is also with School of Automation, Nanjing University of Information Science and Technology, Nanjing 210044, China (e-mail: meyang@connect.hku.hk).

M. Y. Wang is with the Department of Mechanical and Aerospace Engineering and the Department of Electronic and Computer Engineering, Hong Kong University of Science and Technology, Hong Kong (tel.: +852-34692544; e-mail: mywang@ust.hk).

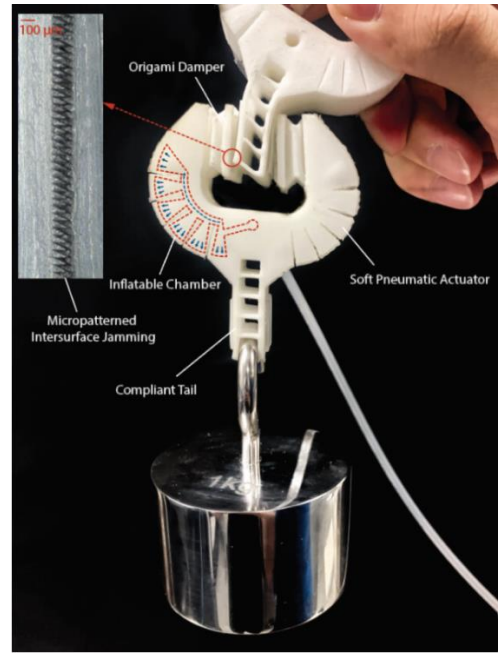

Fig. 1. Two proposed jamming-based flexible connectors able to hold a 1 kg weight under deformation with a detailed view of micropatterned intersurface jamming between two pieces of dry adhesives.

flexible connectors is to integrate electroadhesion pads into soft modules [8]. However, electroadhesion can only support a light weight and needs continuous external power supply.

We contend that a reversible connection mechanism can be integrated into soft structures without impairing the softness while being robust under low power distribution. To this end, we have developed a flexible connector (Fig. 1) for soft modular robots based on intersurface jamming effect of micropatterned dry adhesives. A thin layer of skin-like material with anisotropic surface micropatterns was fabricated by silicone rubber out of a micromachined mold. Engaging two pieces of micropatterned dry adhesives in the opposite direction generates normal and shear adhesion through van der Waals force assisted intersurface jamming, establishing a robust connection. For demonstration, we 3D printed compliant modules with thermoplastic elastomer (TPE) to mount, engage and disengage the micropatterned dry adhesives. Each compliant module has a sunken terminal and a protruding terminal that are compatible; thus multiple modules can be connected end to end.

Through connection strength characterization experiments, we demonstrated how micropatterned intersurface jamming could be implemented as a robust connection mechanism while being fully integrated into compliant structures. After actuation, the mechanism was able to maintain robust connection with small energy consumption.

We also showed the adaptability of the micropattern jamming based connectors by measuring the connection strength of two compliant modules when subjected to disruptions such as misalignment and inadequate contact. The connectors showed passive alignment property and remain robust under large contact lose and deformation.

## II. BACKGROUND

Connection mechanisms for modular robots heretofore can be generally categorized into magnets and mechanical latching. By far, mechanical connectors based on latching are the most popular solution since it oftentimes guarantees robust connection while requiring little energy input. Many modular robots have been successfully made implementing the latching approach [9-18]. A good example is Mori [18], a modular origami robot that utilizes coupling mechanisms to create a strong connection. Robertson et al. presented a modularized soft robotic actuator that is connected with well aligned mechanical couplings [19]. However, mechanical latching or coupling requires solid constraints with additional actuating mechanisms that demands precise alignments and SMA or motor actuation. Therefore, it is unpractical to apply mechanical latching connectors to soft modular robots since it limits the system's overall flexibility.

Another common approach is to use magnetic force to form solid connections between modules while promoting easy attachment. Magnetic connectors can either be static or active depending on the use of permanent magnets or electromagnets, both being used in many successful demonstrations [20-23]. Romanishin et al. demonstrated a momentum-driven, magnetic modular robot that bonds through permanent magnets [24]. Each module can engage and disengage by a torque generated by rapidly decelerating an internal flywheel. Vergara et al. has presented soft modular robotic cubes that can reproduce morphogenetic movements of the Embryo [25]. The modules are connected by permanent magnets as a passive docking mechanism. However, both mechanisms add extra stiffness and weight to the system thus are not ideal for an entirely flexible soft modular system. Other methods such as electroadhesion have also been presented recently as a soft active connection mechanism [26], [27], but it has small connection strength while requiring continuous power supply.

Recently, micropatterned dry adhesives made by soft materials have been integrated into robotic grippers of various stiffness for friction and adhesion enhancement [28-30]. Highly repeatable and robust microzip fastener based on the van der Waals force-assisted interlocking between two micropatterned dry adhesives has also been reported by several groups [31-34]. The micropatterned dry adhesives need little initial pressure to engage and maintain strong bonds without extra pressuring. These works shed new lights on flexible connectors and open up possibilities to create entirely soft yet robust connection mechanisms.

## III. DESIGN AND FABRICATION

### A. Design Overview

Inspired by the interlocking effect between two micropatterned dry adhesives, we designed a flexible connector that composed of flexible pneumatic actuators and micropatterned dry adhesives. The main body of the connector presented in this work are made by flexible TPE, but the design can be further applied to materials with arbitrary stiffness like silicone elastomers. The inflatable chambers can be pneumatically actuated to engage and disengage the pre-mounted adhesive of one connector with the opposed adhesive of another connector. Ideally, the dry adhesives would generate the most jamming adhesion when the contact area is maximized, meaning the two jamming surfaces are expected to always comply with each other. Therefore, the mechanism should also allow full engagement even under deformation.

In the cases where a strong attachment is not needed, for example, during self-reconfiguration, the connector is supposed to be in its neutral state, where the dry adhesives don't accidentally attach to undesired objects. To fulfil this requirement, the dry adhesives should have anisotropic microstructures on one surface; thus intersurface jamming is only effective in the selected direction.

Considering the restrained weight carrying capacity and mobility of typical soft robots, we used light weight TPE and silicone rubber to fabricate the connectors. The choice of materials also allows fast prototyping by 3D printing and simple integration into soft robotic systems directly during casting.

### B. Micropatterned Dry Adhesives

Drawing inspiration from nature, micropatterned dry adhesives are promising candidates for reversible and controllable attachments [35]. The mechanical properties of the adhesives are the dominant factors for attachment characteristics. As a result, the material property and the contact geometry design are the most crucial factors to dry adhesive performances. Based on the aforementioned requirements, we established two basic design principles as a guideline for future development as follow:

1) The dry adhesive should be directional during jamming and only attach to the selected surfaces,

2) The dry adhesive must be completely soft both by nature and during jamming.

Following these design principles, we seek to create a thin layer of anisotropic micropatterned adhesives with silicone elastomer (Fig .2). Based on contact splitting effect, a small feature scale benefits high levels of adhesion [36]; thus the micropatterned structures are at the micron scale. To maximize the contact area, the initial design was to build densely packed wall structured micropatterns. A strong interlocking effect between wall shaped micropatterned surfaces has already been reported in [31]. We further improved the structure with a densely packed tilted V-grooves design that offers a larger contact area and adds the directional feature. As suggested by some articles, a high aspect ratio is desired for optimum adhesion force [37-41]. However, high aspect ratio (AR) structures lack robustness thus prevents reversible adhesion [40], [42]. Therefore, for the final design, we used a low aspect ratio (AR~2) and a feature scale (depth of the V-grooves) of 100 μm (Fig. 2b).

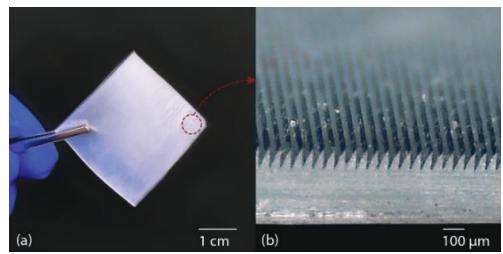

Fig. 2. A piece of micropatterned dry adhesive pad with anisotropic micro V-groove surface structures. (a) The overview of a 3 × 3 cm micropatterned dry adhesive pad. (b) Detailed view of the anisotropic micro V-groove patterns.

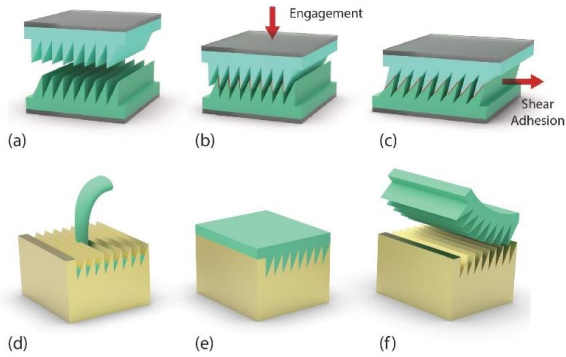

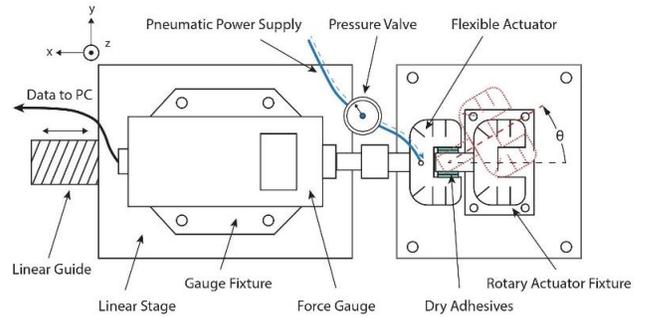

Fig. 3. Micropatterned intersurface jamming and dry adhesives fabrication process flows. (a-c) An illustration of the engagement process of micropatterned intersurface jamming. (a) Initial state where no contact occurs. (b) Engagement of two dry adhesive pads to initiate contact. (c) Drag in shear direction increases the contact area to the maximum. (d-f) Fabrication procedures of micropatterned dry adhesives. (d) Liquid silicone elastomers poured into micromachined surface patterned molds. (e) Cure for 15 hours. (f) Demolding.

Fig. 4. Schematic diagram of the experimental setup for connection force testing on both linear and rotational direction

Initially, the jamming effect is off since there is zero contact (Fig. 3a). When two pieces of adhesives are engaged with each other in the opposed direction (Fig. 3b), the microstructures enter each other's cavities to form interlocking and jamming starts. The intersurface jamming force is expected to be the strongest in the opposed shear direction (Fig. 3c), where the most contact area is presented. Therefore, the jamming effect between micropatterned dry adhesives can be directional, thus prevents them from attaching accidentally.

The micropatterned dry adhesives are fabricated by casting SortaClear 40 silicone elastomers (Smoothon) onto a micromachined surface mold (Fig. 3d). The metal mold is machined by ultraprecision diamond ruling that produces an optical level surface finish. A smooth surface finish is crucial to form strong interlocking effect since it produces the maximum contact surface area. The adhesive is allowed a 24 h cure before demolding (Fig. 3e). The micro V-groove features from the mold are then replicated perfectly to the silicone surface (Fig. 3f), and the adhesive is ready to use. The thickness of the dry adhesives can be reduced to human skin level of several hundred microns; thus it can be attached to any device without adding extra weight or volume.

### C. Active Engage and Disengage Mechanism

The micropatterned dry adhesives have the maximum intersurface jamming force when the adhesive pads are fully engaged and loaded in the shear direction. Therefore, the principle of the connector's engagement mechanism design is to make the engaged dry adhesives parallel when initially loaded. Conventional parallel gripper designs cannot achieve this functionality because when using soft materials, the two fingertips will easily expand and cause stress singularities on the gecko pads.

To address the issue, two key factors are taken into consideration. On the one hand, the initial state of the connectors when loaded, should give a gradient stress distribution on the dry adhesives with higher stress at the distal end and lower stress at the proximal end. On the other hand, the contact area between two connectors should cover as large proportion of the dry adhesives as possible when loaded. As a result, we can take full advantage of the dry adhesives and maximize the intersurface jamming force.

Generally, a single connector can be decomposed into three components (Fig. 1). One component is the main body actuator, which is essentially a soft pneumatic gripper with inflatable chambers. The pneumatically actuated main body can engage and disengage the dry adhesives on two connectors and adjust the connection force through chamber inflation pressure. Another component is the two origami-inspired dampers on both fingertips. The two dampers are basically individual chambers that are not connected to a pneumatic power source and sever only as passive alignment mechanisms for maximizing the contact area of the dry adhesives when the actuators are inflated. In other words, the two dampers generate the gradient stress distribution we mentioned on the dry adhesives to avoid insufficient contact during loading.

The last component is the compliant connector tail for connection to another connector unit. The compliant tail is ladder-shaped, thus it can deform to change orientation when subjected to external disturbance force or environmental constraints. Moreover, the design is also inspired by the fin-ray structures which are compatible with different shapes. The compliant ladder-shape tail presented here is expected to maintain large contact area between dry adhesives even when the structure is highly deformed.

## IV. CHARACTERIZATION

### A. Connection Force Characterization

To characterize the performance of the flexible connector, two experiments were performed. The first experiment aims to evaluate the connection force between two connectors under different chamber inflation pressure. The connection force test was conducted on a custom-built apparatus (Fig. 4). Before measurement, two connectors were mounted onto pre-aligned fixtures to make sure they are coaxial. One of the fixtures was mounted firmly on a platform, while the other fixture was connected to a digital force gauge (range: 0>30 N, precision: 0.01 N) mounted on a linear guide. The actuators are then inflated by a pressure $P_{inflation}$ to engage the adhesives. During the test, the force gauge can move horizontally along the linear guide and record the shear force profiles between the connectors through time until detachment. A total of six inflation pressure was tested from 0 to 2.8 bar, with each case repeated for five times. Four pieces of 1.5 × 1.5 cm were used in pairs throughout the entire experiments, which gives a constant contact area of 4.5 cm².

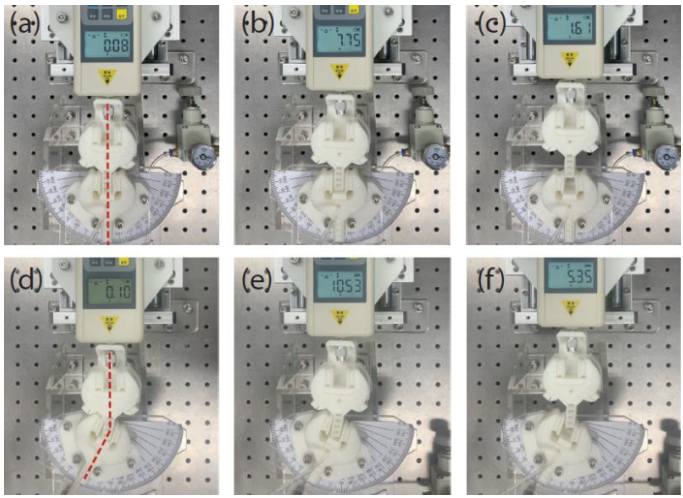

Fig. 5. Experiment process for linear (a-c) and rotational (d-f) direction. (a) Linear direction force test initial state. (b) Maximum connection force state. (c) Collapse state. (d) 30º planar rotational direction force test initial state. (e) Maximum connection force state. (f) Collapse state.

The force and time relationships for the six inflation pressures are illustrated in Fig. 6a. The plot was generated using the set of data that shows the maximum shear force out of the four trials for each pressure setting. The measured maximum and average maximum connection force at six different $P_{inflation}$ are listed in Table. 1.

As shown in Fig. 6a, dry adhesives under all inflation pressures experienced a low initial connection force followed by a drastic increase and ended with a gradual decrease until break. Initially, the dry adhesives are not in full contact due to misalignments thus the jamming force is low. Afterwards, the pressure chambers were inflated to engage the dry adhesives and eliminated the misalignments (Fig. 5a). When dragged in the shear direction (Fig. 5b), the slanted faces of the V-grooves were brought into contact instead of the vertical faces (Fig. 3c), further increasing the contact area and resulted in a drastic increase of the connection force. After a maximum jamming force was achieved, the dry adhesives started sliding out contact and gradually disconnected until the connection breaks (Fig. 5c). From our measurements, a maximum connection force of 23.52 N was recorded at an inflation pressure of 2.8 bar. As expected, the maximum connection force increased with the $P_{Inflation}$. Surprisingly, an average maximum connection force of 1.75 N was recorded at zero inflation pressure. This result indicates that intersurface jamming between dry adhesives is effective as long as contact occurs, even with minimum preload pressure in the normal direction. We argue that such characteristic is especially beneficial when high pressure is unavailable, for example, in the case where an ultra-soft main body is presented.

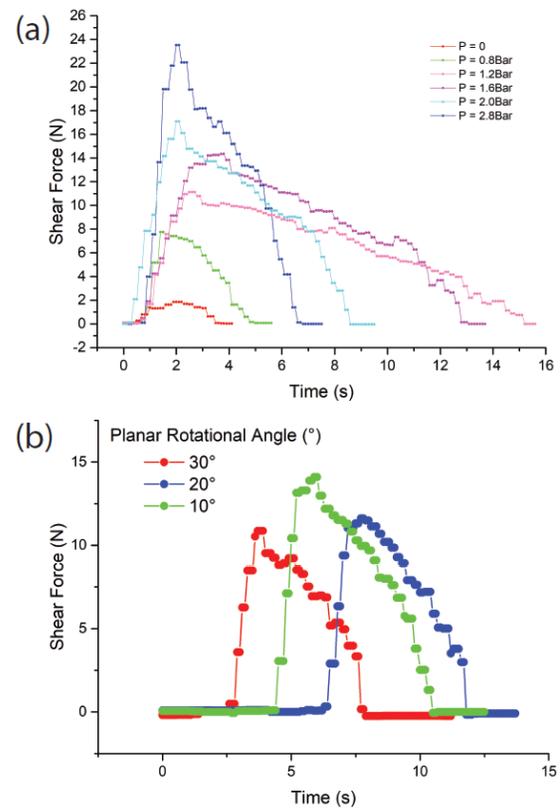

Fig. 6. Force-time curve from connection force tests. (a) Linear direction at 6 inflation pressure. (b) Planar rotational direction at 2 bar inflation pressure.

To find out the connector performance under planar rotational distortion, we measured the connection force at three rotational angles in the second experiment. Using a similar experimental setup as the first experiment (Fig. 4), we rotated one actuator by 10, 20 and 30 degrees respectively and fixed it for measurement. The experimental setup for the connection force test at the 30º rotational direction is shown by Fig. 5d-f as an example. Measurement at each angle was repeated for three times at the same inflation pressure of 2 bar and the results are plotted in Fig. 6b using the best set of data.

We found that although a planer rotation of 30 degrees (Fig. 5d-f) resulted in a drop of maximum connection force from 17 N to 9.7 N, the remaining connection force is still large enough to maintain a robust connection. The drop in connection force is mainly due to the loss of contact area and the peel-off effect caused by stress singularities. When a planer rotation is presented (Fig. 5d), high inflation pressure is necessary to align the dry adhesives. As a result, the stress concentration is not equal with a high singularity at the initial contact point. When dragged along the shear direction, the flexible connectors first complied with each other and the contact area increased (Fig. 5e). After the maximum connection force was achieved, a rapid detachment was observed (Fig. 5f), thus the connection quickly collapsed.

TABLE I. LINEAR CONNECTION FORCE TEST RESULTS

| $P_{Inflation}$ (bar) | 0 | 0.8 | 1.2 | 1.6 | 2.0 | 2.8 |
|---|---|---|---|---|---|---|
| Maximum Force (N) | 1.84 | 7.75 | 11.12 | 14.35 | 17.09 | 23.52 |
| Average Maximum Force (N) | 1.75 | 7.14 | 10.87 | 14.00 | 16.40 | 22.31 |

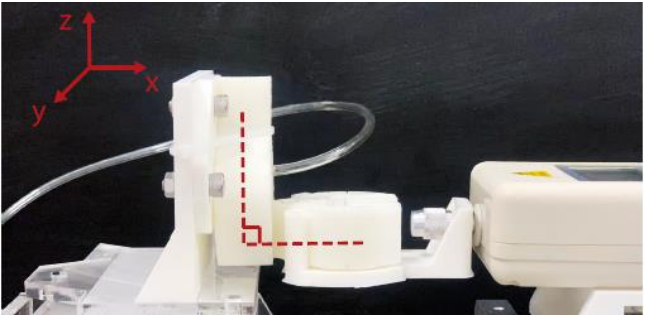

Fig. 7. Experimental setup of connection force tests at 90º rotation around y-axis. Note that the 30º and 60º tests were performed using tilted fixtures on the same platform.

### B. Selective Attachment

Previous experiments focused on testing the connector's performance on the planar level. To expand our understanding, we measured the connection force when the actuator is rotated along the y-axis (Fig. 7). A total of three different angles were tested (30 °, 60º and 90º) at different inflation pressure (0, 1.2 and 2.0 bar). We found around 50% loss in connection force for all three angles at high inflation pressure compared to the shear direction (Fig. 8a). This can be explained by the reduction in van der Waals force-assisted intersurface jamming between dry adhesives. When tilted around y-axis, the micro V-grooves became misaligned, thus intersurface jamming is prohibited even though the dry adhesives are in perfect contact. As a result, the micropatterned dry adhesives behaved similarly to smooth silicone rubber pads. This can be further confirmed by the results in the next section (Fig. 8b), where we replaced the micropatterned dry adhesives with smooth silicone rubber pads of the same material. The connection force of the two cases were almost identical. From the results, the connectors were confirmed to show selectivity in attachment direction with the strongest connection force in the shear direction.

### C. Comparative Tests

To further illustrate the advantage of jamming based connecters, we replaced the micro patterned dry adhesives with smooth silicone pads and pure TPE respectively. The silicone rubber pads were made with the same material of the dry adhesives. Additionality, the thickness of the silicone pads was controlled to be the same with the dry adhesives. The results showed that the jamming-based connector has a more than 50% increase in connection force than the ones with silicone pads and is almost ten times stronger than the pure TPE connectors without dry adhesives, as shown in Fig. 8b. In addition, the slope of the force and pressure curve of the jamming-based connector was the highest, which means it has better pneumatic energy to connection strength transfer rate.

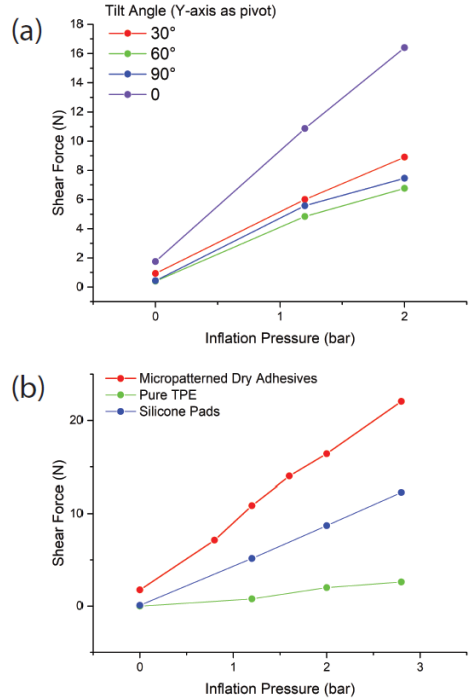

Fig. 8. Comparison of maximum shear force at different inflation pressure. (a) Comparison among four tilt angles along y-axis (0º - 90º) at 0, 1.2 and 2.0 bar. (b) Comparison between micropatterned dry adhesives, pure TPE and smooth silicone pads.

## V. CONCLUSION AND FUTURE WORK

Our work presents a first study in design, fabricating and characterzing a flexible connector for soft modular robots based on micropatterned surface jamming. The proposed connector has micropatterned dry adhesives made by silicone rubber for intersurface jamming. When two pieces of dry adhesives engaged and dragged along the shear direction, a robust connection can be formed. The connector also has a flexible, 3D printed TPE main body with inflatable chambers for active engagement and disengagement. Origami dampers on the main body enable passive alignment and maximize the contact area of the dry adhesives under deformation. The overall stiffness of the connector is low; thus it can be integrated into soft modular robot bodies without impairing the flexibility of the system.

In this paper, we characterized the connection force of the proposed connector. The connector can support a maximum linear load of 23.52 N and a 10.86 N load under 30º rotation. The connector also show selective attachment properties where the intersurface jamming effect is only strong at the specified direction. Additionally, the jamming-based connector is more than twice as strong as connectors using pure silicone pads and ten times stronger than the ones using pure TPE. Last but not least, the connectors are proofed to be more energy efficient.

The immediate pursue of the future study is to improve the performance of the jamming-based connectors under low inflation pressure. We will also further implement the connectors to build entirely soft modular robots. Compensating strategies of the connectors under real world disturbances can then be studied.